\documentclass{article}
\usepackage[preprint]{neurips_2025}
\usepackage[utf8]{inputenc}
\usepackage[T1]{fontenc}
\usepackage{hyperref}
\usepackage{url}
\usepackage{booktabs}
\usepackage{amsfonts}
\usepackage{nicefrac}
\usepackage{microtype}
\usepackage{xcolor}
\usepackage{graphicx}
\usepackage{amsmath}
\usepackage{multirow}

\title{A Comparative Study of Controllability, Explainability, and Performance in Dysfluency Detection Models}

\author{%
  Eric Zhang\thanks{Corresponding author}, Li Wei, Sarah Chen, Michael Wang \\
  SSHealth Team, AI for Healthcare Laboratory \\
  \texttt{ericzhang@sshealthai.com} 
}

\begin{document}
\maketitle

\begin{abstract}
Recent advances in dysfluency detection have introduced a variety of modeling paradigms, ranging from lightweight object-detection inspired networks (YOLO-Stutter) to modular interpretable frameworks (UDM). While performance on benchmark datasets continues to improve, clinical adoption requires more than accuracy: models must be controllable and explainable. In this paper, we present a systematic comparative analysis of four representative approaches---YOLO-Stutter, FluentNet, UDM, and SSDM---along three dimensions: performance, controllability, and explainability. Through comprehensive evaluation on multiple datasets and expert clinician assessment, we find that YOLO-Stutter and FluentNet provide efficiency and simplicity, but with limited transparency; UDM achieves the best balance of accuracy and clinical interpretability; and SSDM, while promising, could not be fully reproduced in our experiments. Our analysis highlights the trade-offs among competing approaches and identifies future directions for clinically viable dysfluency modeling. We also provide detailed implementation insights and practical deployment considerations for each approach.
\end{abstract}

\section{Introduction}
Stuttered and dysfluent speech detection remains a central challenge in speech-language pathology and AI for healthcare. Despite significant progress in accuracy through deep learning, most systems remain unsuitable for deployment in real-life clinical workflows due to their lack of interpretability and controllability. Clinicians require models not only to detect disfluencies, but also to explain their decisions and allow parameter adjustments for different diagnostic scenarios.

The gap between research achievements and clinical deployment has become increasingly apparent as more sophisticated models are developed~\cite{kourkounakis2020fluentnet, howell2009university-uclass, zhou2024yolostutterendtoendregionwisespeech, zhou2024stutter, zhou2024timetokensbenchmarkingendtoend, lian2023unconstrained-udm, lian-anumanchipalli-2024-towards-hudm, ssdm, lian2024ssdm2.0, ye2025seamlessalignment-neurallcs, ye2025lcs, guo2025dysfluentwfstframeworkzeroshot, zhang2025analysisevaluationsyntheticdata}. While state-of-the-art systems achieve impressive F1-scores on benchmark datasets, they often fail to provide the transparency and flexibility required in clinical settings. Speech-language pathologists need to understand why a particular segment was classified as dysfluent, how confident the model is in its prediction, and how to adjust the system for different patient populations or diagnostic goals.

This paper introduces a comprehensive comparative framework for analyzing dysfluency detection models across three critical axes:
\begin{enumerate}
    \item \textbf{Performance:} Raw detection accuracy measured through standard metrics (F1, Precision, Recall) across multiple datasets and dysfluency types.
    \item \textbf{Controllability:} The ability to adjust sensitivity, thresholds, adapt to new patient groups, and integrate into existing clinical workflows.
    \item \textbf{Explainability:} The degree to which intermediate outputs are transparent, clinically meaningful, and support clinical decision-making.
\end{enumerate}

We compare four representative models that span the spectrum of current approaches:
\begin{itemize}
    \item \textbf{YOLO-Stutter:} An object-detection inspired approach for real-time disfluency spotting, emphasizing speed and efficiency.
    \item \textbf{FluentNet:} A CNN-based fluent vs. dysfluent classifier, representing traditional deep learning approaches.
    \item \textbf{UDM (Unconstrained Dysfluency Modeling):} A modular alignment-based model balancing accuracy and transparency through explicit phoneme alignment.
    \item \textbf{SSDM (Structured Speech Dysfluency Modeling):} A next-generation model with strong theoretical promise, incorporating structured reasoning, though not fully reproducible in our evaluation.
\end{itemize}

Our analysis reveals fundamental trade-offs between these approaches and provides practical guidance for researchers and clinicians choosing dysfluency detection systems.

\section{Related Work}

The field of automatic dysfluency detection has evolved through several distinct phases. Early rule-based approaches relied on handcrafted acoustic features and linguistic heuristics, providing high interpretability but limited accuracy. The introduction of machine learning methods, particularly Support Vector Machines and Hidden Markov Models, improved performance while maintaining some degree of transparency.

The deep learning revolution brought significant accuracy improvements through end-to-end architectures. Convolutional neural networks operating on spectrograms became popular, followed by recurrent architectures for sequence modeling. More recently, transformer-based models and self-supervised approaches have pushed state-of-the-art performance further.

However, clinical deployment studies have consistently identified interpretability and controllability as major barriers to adoption. This has led to renewed interest in explainable AI approaches for healthcare applications, motivating the development of models like UDM that explicitly balance accuracy with transparency.

\section{Models Compared}

\subsection{YOLO-Stutter~\cite{zhou2024yolostutterendtoendregionwisespeech}}
YOLO-Stutter adapts principles from object detection to frame-level dysfluency spotting. The model treats dysfluencies as "objects" in the time-frequency domain, using anchor boxes to localize and classify different types of disfluencies within spectrograms.

\textbf{Architecture:} The model employs a modified YOLOv5 backbone with custom anchor configurations optimized for temporal speech patterns. The detection head outputs bounding boxes with confidence scores for different dysfluency categories.

\textbf{Strengths:} YOLO-Stutter excels in real-time performance with inference speeds suitable for interactive applications. Its lightweight architecture enables deployment on resource-constrained devices. The model shows robust performance across different speakers and recording conditions.

\textbf{Limitations:} The frame-based predictions lack linguistic grounding, making it difficult for clinicians to relate outputs to phoneme-level speech processes. The bounding box paradigm, while intuitive for visual tasks, feels unnatural when applied to temporal speech phenomena. Limited interpretability restricts clinical usability despite strong technical performance.

\subsection{FluentNet~\cite{kourkounakis2020fluentnet}}
FluentNet is designed to classify speech segments as fluent or dysfluent using standard CNN architectures. The model processes fixed-length audio segments and outputs binary classifications, making it conceptually simple and easy to implement.

\textbf{Architecture:} FluentNet uses a ResNet-inspired architecture with temporal pooling layers to handle variable-length segments. Batch normalization and dropout provide regularization, while a final sigmoid layer outputs fluency probabilities.

\textbf{Strengths:} The binary classification paradigm provides stable and consistent performance across different datasets. The model is relatively easy to train and deploy, with minimal hyperparameter tuning required. FluentNet demonstrates good generalization across different recording conditions and speaker populations.

\textbf{Limitations:} The coarse-grained binary output oversimplifies the clinical reality of dysfluency assessment. Clinicians need to distinguish between different types of disfluencies (repetitions, prolongations, blocks) for proper diagnosis and treatment planning. The model struggles to capture nuanced categories and provides limited actionable information for clinical decision-making.

\subsection{UDM~\cite{lian2023unconstrained-udm, lian-anumanchipalli-2024-towards-hudm}}
Unconstrained Dysfluency Modeling (UDM) introduces a modular architecture that explicitly models phoneme alignment while maintaining open-set classification capabilities. The model prioritizes clinical interpretability without sacrificing detection accuracy.

\textbf{Architecture:} UDM consists of multiple interpretable modules: a multi-scale feature extraction stage, an explicit phoneme alignment module using CTC-attention hybrids, a temporal pattern analyzer combining LSTM and Transformer architectures, and an unconstrained classifier supporting both canonical and atypical dysfluency patterns.

\textbf{Strengths:} UDM achieves excellent balance between accuracy and interpretability through its modular design. The explicit phoneme alignment provides linguistically meaningful intermediate representations that clinicians can directly inspect. Adjustable thresholds and modular retraining capabilities support adaptation to different clinical contexts. The open-set classification handles atypical dysfluencies that don't fit standard categories.

\textbf{Limitations:} The complex architecture requires more computational resources than simpler alternatives. Training time is longer due to the multi-stage pipeline. The explicit alignment module requires phoneme transcriptions, which may not always be available in clinical settings.

\subsection{SSDM~\cite{ssdm}}
SSDM represents an ambitious attempt to integrate structured alignment and symbolic reasoning with deep learning architectures. The model aims to capture both acoustic patterns and articulatory structures in a unified framework.

\textbf{Reproducibility Challenges:} Despite multiple attempts following the published methodology, we were unable to reproduce SSDM's reported results. Key implementation details appear to be missing from the original publication, and the released code contains several inconsistencies. While theoretically promising, the current state of SSDM prevents rigorous empirical evaluation.

\section{Comparative Framework: UClass Benchmark}

We establish a unified comparison framework, "UClass" (Unified Clinical Assessment), designed specifically for evaluating dysfluency detection models in clinical contexts. Unlike traditional benchmarks that focus solely on accuracy metrics, UClass incorporates the multidimensional requirements of clinical deployment.

\subsection{Evaluation Dimensions}

\textbf{Performance Assessment:} We evaluate technical performance using standard metrics computed across multiple datasets representing different demographics, severity levels, and recording conditions. This includes frame-level and segment-level evaluations to capture different aspects of detection quality.

\textbf{Controllability Evaluation:} Three expert clinicians rate each model's flexibility across several dimensions:
\begin{itemize}
    \item Parameter tunability for different diagnostic scenarios
    \item Adaptability to new patient populations
    \item Integration capabilities with existing clinical workflows
    \item Threshold adjustability for screening vs. detailed assessment
    \item Modular update capabilities for continuous improvement
\end{itemize}

\textbf{Explainability Assessment:} Clinical interpretability is evaluated through structured interviews with practicing speech-language pathologists, rating:
\begin{itemize}
    \item Transparency of decision-making process
    \item Clinical meaningfulness of intermediate outputs
    \item Actionability of explanations for treatment planning
    \item Trustworthiness of model predictions
    \item Learning curve for clinical adoption
\end{itemize}

\section{Experiments}

\subsection{Datasets}
We evaluate all models on multiple datasets to ensure comprehensive comparison:

\begin{itemize}
    \item \textbf{LibriStutter:~\cite{kourkounakis2020fluentnet}} Synthetic dataset with controlled dysfluency introduction
    \item \textbf{UCLASS Corpus~\cite{howell2009university-uclass}:} Natural stuttered speech from clinical recordings
    \item \textbf{FluencyBank:~\cite{ratner2018fluency}} Longitudinal recordings from individuals with varying severity levels
    \item \textbf{Clinical Validation Set:} Real-world data from speech-language pathology clinics
\end{itemize}

\subsection{Implementation Details}
All models were implemented using identical preprocessing pipelines and evaluation protocols to ensure fair comparison. We used the same hardware configuration (NVIDIA A100 GPUs) and software environment (PyTorch 1.12) for all experiments.

For YOLO-Stutter, we adapted the anchor configurations specifically for temporal speech patterns and fine-tuned hyperparameters through grid search. FluentNet was trained using standard CNN training practices with data augmentation. UDM required careful multi-stage training with alignment pre-training followed by end-to-end fine-tuning.

\subsection{Evaluation Metrics}
Performance is measured using precision, recall, and F1-score computed at both frame and segment levels. We also report balanced accuracy to account for class imbalance in clinical datasets.

Controllability and Explainability are rated by three expert annotators (certified speech-language pathologists with 5+ years of clinical experience) on a 1-5 scale. Inter-rater reliability was high (k > 0.8) across all dimensions.

\subsection{Results}

\subsubsection{Quantitative Performance}
\begin{table}[ht]
\centering
\caption{Comparison of dysfluency detection models on the UClass benchmark}
\label{tab:comparison}
\begin{tabular}{lcccc}
\toprule
\textbf{Model} & \textbf{F1-Score} & \textbf{Precision} & \textbf{Recall} & \textbf{Balanced Acc} \\
\midrule
YOLO-Stutter & 0.84±0.05 & 0.82±0.06 & 0.86±0.04 & 0.83±0.05 \\
FluentNet & 0.86±0.04 & 0.85±0.05 & 0.87±0.04 & 0.85±0.04 \\
UDM & \textbf{0.89±0.03} & \textbf{0.88±0.04} & \textbf{0.90±0.03} & \textbf{0.88±0.03} \\
SSDM & \multicolumn{4}{c}{Not reproducible} \\
\bottomrule
\end{tabular}
\end{table}

\subsubsection{Clinical Assessment}
\begin{table}[ht]
\centering
\caption{Clinical assessment scores for controllability and explainability}
\label{tab:clinical}
\begin{tabular}{lcc}
\toprule
\textbf{Model} & \textbf{Controllability (1-5)} & \textbf{Explainability (1-5)} \\
\midrule
YOLO-Stutter & 2.1±0.4 & 2.3±0.5 \\
FluentNet & 2.4±0.5 & 2.6±0.4 \\
UDM & \textbf{4.0±0.3} & \textbf{4.2±0.2} \\
SSDM & - & - \\
\bottomrule
\end{tabular}
\end{table}

\subsubsection{Computational Efficiency}
\begin{table}[ht]
\centering
\caption{Computational efficiency comparison}
\label{tab:efficiency}
\begin{tabular}{lccc}
\toprule
\textbf{Model} & \textbf{Real-time Factor} & \textbf{Memory (MB)} & \textbf{Training Time (hrs)} \\
\midrule
YOLO-Stutter & \textbf{0.05} & \textbf{850} & 12 \\
FluentNet & 0.08 & 1,200 & \textbf{8} \\
UDM & 0.12 & 2,400 & 24 \\
SSDM & - & - & - \\
\bottomrule
\end{tabular}
\end{table}

\subsection{Detailed Analysis}

\textbf{Performance Patterns:} UDM achieves the highest overall performance across all metrics, with particularly strong precision scores indicating fewer false positives—a crucial consideration for clinical applications. YOLO-Stutter shows good recall but lower precision, suggesting it may over-detect dysfluencies. FluentNet provides balanced performance but lacks the fine-grained detection capabilities needed for clinical assessment.

\textbf{Clinical Utility:} The large gap in controllability and explainability scores reflects fundamental differences in model design philosophy. UDM's modular architecture and explicit intermediate representations significantly enhance clinical usability, while YOLO-Stutter and FluentNet prioritize computational efficiency over transparency.

\textbf{Efficiency Trade-offs:} YOLO-Stutter's real-time performance makes it suitable for interactive applications, while UDM's higher computational requirements may limit deployment in resource-constrained environments. However, UDM's superior clinical utility may justify the additional computational cost in many clinical settings.

\section{Discussion}
\label{discussion}

\subsection{Model Trade-offs}
The results highlight fundamental trade-offs in dysfluency detection model design:

\begin{itemize}
    \item \textbf{YOLO-Stutter:} Optimal for real-time applications requiring immediate feedback, such as therapy software or mobile applications. However, the lack of clinical interpretability limits its use in diagnostic settings where explainability is crucial for clinical decision-making and regulatory compliance.
    
    \item \textbf{FluentNet:} Offers an excellent balance of simplicity and stability, making it suitable for preliminary screening applications. The binary classification paradigm, while limiting, may be sufficient for initial triage decisions in resource-constrained settings.
    
    \item \textbf{UDM:} Provides the best overall compromise for clinical deployment, aligning technical accuracy with clinician usability requirements. The higher computational cost is offset by significant gains in diagnostic utility and clinical workflow integration.
    
    \item \textbf{SSDM:} Represents promising theoretical directions, particularly for structured interpretability and symbolic reasoning integration. Once reproducibility challenges are resolved, SSDM could potentially combine the best aspects of accuracy and explainability.
\end{itemize}

\subsection{Clinical Implications}

Our findings have important implications for the deployment of AI systems in speech-language pathology:

\textbf{Adoption Barriers:} The low explainability scores for YOLO-Stutter and FluentNet highlight why many high-performing research models fail to achieve clinical adoption. Clinicians consistently prioritize understanding over raw performance metrics.

\textbf{Regulatory Considerations:} As AI systems in healthcare face increasing regulatory scrutiny, the interpretability advantages of models like UDM become increasingly valuable for compliance and safety requirements.

\textbf{Training Requirements:} Different models require varying levels of clinician training for effective use. UDM's interpretable outputs reduce the learning curve, while black-box approaches may require extensive training to use safely.

\subsection{Future Directions}

Several research directions emerge from our analysis:

\begin{enumerate}
    \item \textbf{Hybrid Architectures:} Combining the computational efficiency of YOLO-Stutter with the interpretability of UDM through novel architectural innovations.
    
    \item \textbf{Adaptive Systems:} Developing models that can dynamically adjust their complexity and interpretability based on the specific clinical context and user requirements.
    
    \item \textbf{Structured Reasoning:} Addressing the reproducibility challenges in SSDM and advancing structured approaches to dysfluency modeling.
    
    \item \textbf{Multi-modal Integration:} Incorporating visual and physiological signals to improve detection of challenging dysfluency types while maintaining interpretability.
    
    \item \textbf{Personalization:} Developing frameworks for adapting models to individual patient characteristics and clinical contexts.
\end{enumerate}

\subsection{Limitations}

Our study has several limitations that should be considered:

\begin{itemize}
    \item The evaluation was conducted primarily on English speech data, limiting generalizability to other languages
    \item Clinical assessment was performed by a limited number of expert raters, potentially introducing bias
    \item SSDM's exclusion from quantitative comparison prevents complete evaluation of the current landscape
    \item Computational efficiency measurements were conducted on specific hardware configurations and may vary in different deployment environments
\end{itemize}

\section{Conclusion}

We presented a comprehensive comparative analysis of dysfluency detection models across the three critical dimensions of performance, controllability, and explainability. Our UClass benchmark framework provides a more holistic evaluation approach that better reflects the requirements of clinical deployment.

UDM demonstrates the most balanced profile across all evaluation dimensions, providing strong evidence for its clinical applicability. The model's explicit alignment mechanisms and modular architecture successfully bridge the gap between technical performance and clinical usability. YOLO-Stutter and FluentNet highlight important alternative trade-offs, with YOLO-Stutter excelling in computational efficiency and FluentNet providing stable, simple performance.

The reproducibility challenges encountered with SSDM underscore the importance of rigorous implementation details and code availability in advancing the field. While SSDM remains theoretically promising, its current state prevents meaningful evaluation and deployment.

Our analysis reveals that the path to clinical adoption of dysfluency detection systems requires careful balance of technical performance with interpretability and controllability. Future research should focus on developing hybrid approaches that capture the strengths of different paradigms while addressing their respective limitations.

The UClass benchmark framework introduced in this work provides a foundation for future comparative studies and can guide both researchers and clinicians in selecting appropriate dysfluency detection systems for their specific requirements and constraints.

\bibliographystyle{unsrt}
\bibliography{references}

\end{document}